\documentclass{article}
\usepackage{engram_preprint,times}
\usepackage[T1]{fontenc}
\usepackage[utf8]{inputenc}
\usepackage{amsmath,amssymb,booktabs,multirow,graphicx}
\usepackage{algorithm,algpseudocode}
\usepackage{microtype}
\usepackage{listings}
\usepackage[table]{xcolor}
\usepackage{tikz}
\usetikzlibrary{arrows.meta,positioning,fit,calc,shapes.geometric}
\usepackage{hyperref}
\usepackage{url}
\hypersetup{
  hidelinks,
  pdftitle={ENGRAM: From Agent Experience to Weights through Verified Consolidation},
  pdfsubject={Research preprint}
}
\newcommand{\method}{\textsc{Engram}}
\newcommand{\memory}{\mathcal{M}}
\newcommand{\replay}{\mathcal{R}}
\newcommand{\skills}{\mathcal{S}}
\newcommand{\dataset}{\mathcal{D}}
\newcommand{\probe}{\mathcal{V}}
\newcommand{\ind}{\mathbb{I}}

\newcommand{\weights}{\mathrm{W}}
\newcommand{\external}{\mathrm{M}}
\newif\ifdraftpaper
\draftpaperfalse
\definecolor{draftgray}{HTML}{666666}
\definecolor{engblue}{HTML}{EAF2FC}
\definecolor{engpurple}{HTML}{F1ECF7}
\definecolor{enggreen}{HTML}{EAF3E8}
\definecolor{engyellow}{HTML}{FFF4DA}
\definecolor{engred}{HTML}{FAEAE8}

\newcommand{\bestresult}[1]{\cellcolor{enggreen}{\bfseries\boldmath #1}}
\newcommand{\secondresult}[1]{\cellcolor{engyellow}{#1}}

\title{ENGRAM: From Agent Experience to Weights\\through Verified Consolidation}
\author{%
  \textbf{Ruiyi Yang\textsuperscript{1}}\quad
  \textbf{Zechen Li\textsuperscript{1}}\quad
  \textbf{Lihuan Li\textsuperscript{1}}\\[2pt]
  \textbf{Du Yin\textsuperscript{1}}\quad
  \textbf{Hao Xue\textsuperscript{2,1}}\quad
  \textbf{Flora D. Salim\textsuperscript{1}}\\[5pt]
  \normalfont\textsuperscript{1}University of New South Wales\\
  \normalfont\textsuperscript{2}The Hong Kong University of Science and Technology (Guangzhou)\\[2pt]
  \normalfont\small\href{mailto:ruiyi.yang@unsw.edu.au}{\texttt{ruiyi.yang@unsw.edu.au}}%
}
\hypersetup{pdfauthor={Ruiyi Yang, Zechen Li, Lihuan Li, Du Yin, Hao Xue, Flora D. Salim}}

\date{}
\begin{document}
\maketitle
\begin{abstract}
Language-model agents use experience through external retrieval and parameter learning. Consolidating stored experience into model weights requires deciding both which experience to train on and when its external records can leave retrieval. Admission based solely on student mastery can exclude useful teacher demonstrations, while removal immediately after training can discard guidance that the model still needs. We introduce \method, a framework that separates training admission from memory release. A demonstration-aware gate admits skills based on student mastery or sufficient successful teacher demonstrations. The student learns from selected action traces with context from an Engram Trace Graph (ETG). A verification gate then compares the updated model without episodic retrieval against the base model with retrieval on reserved probes. Release requires valid training participation and scores meeting both relative and absolute criteria; records for skills that fail verification remain retrievable. Experiments across interactive and reasoning benchmarks show consistent gains over frozen-weight ETG controls, with ablations supporting demonstration-aware admission and selective adapter activation. A controlled held-out comparison shows that verified release substantially reduces the retrievable footprint while remaining close to retain-all performance and outperforming immediate and budget-matched random release. \method\ also improves task performance while reducing input-token usage and model calls. Skill-level verification further reduces retrievable experience through selective archival, retaining external guidance for skills that do not meet the release criteria. Code is available in anonymous.4open.science/r/engram-anonymous.
\end{abstract}

\section{Introduction}
\label{sec:intro}

Language-model agents use past experience through external retrieval and parameter updates. Retrieval keeps trajectories and feedback available for future tasks, while parameter learning can make their lessons usable without consulting the original records. Combining these mechanisms raises a practical question: \emph{when has an agent learned enough from stored experience to remove the corresponding records from its retrieval memory?} Retaining every record incurs storage and retrieval costs, but premature removal can discard guidance that the model still needs.

Existing methods illustrate both approaches. Reflexion stores verbal feedback, Voyager accumulates executable skills, and MAGE organizes successful and failed experiences in a co-evolving graph \citep{shinn2023reflexion,wang2023voyager,yang2026mage}. Self-training methods such as STaR and ReST$^{\mathrm{EM}}$ instead update model parameters using selected generated solutions \citep{zelikman2022star,singh2024restem}. An agent combining retrieval and training must distinguish useful learning material from records that are no longer needed after learning.

\textbf{First, useful training experience may precede student mastery.} A student can fail at prerequisite actions before reaching a more advanced skill. A stronger teacher can demonstrate that skill, but a mastery-only admission rule would still exclude it from training. Admission should therefore consider available demonstrations alongside current competence.

\textbf{Second, completing training does not establish that memory can be removed.} A selected skill may contribute no valid training examples, or the updated model may still depend on retrieved guidance. Premature removal creates \emph{knowledge waste}: external experience becomes unavailable before the weights can replace it. This can happen immediately after consolidation, independently of interference from later updates.

\begin{figure}[t]
\centering
\includegraphics[width=\linewidth]{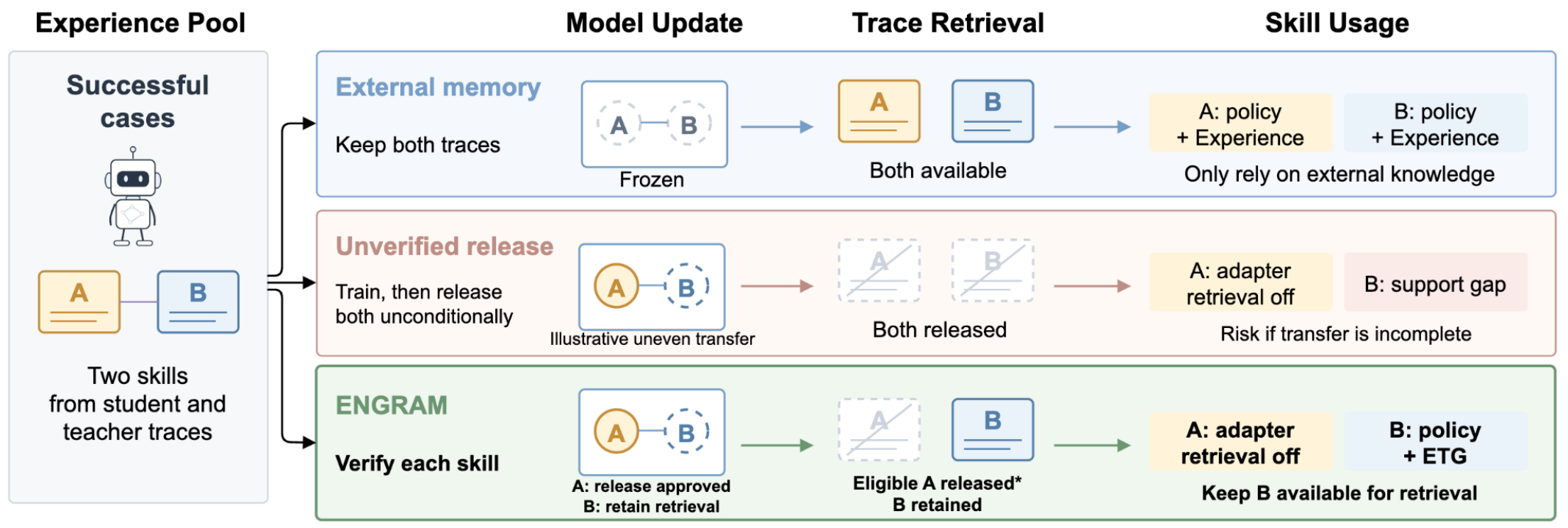}
\caption{\textbf{Readiness to learn differs from readiness to release.} Removing both traces after training can leave skill B unsupported if only A has been learned. \method\ verifies each skill: A uses verified weights and releases eligible records, while B retains external memory. Configured residuals remain; colored nodes represent behavior.}
\label{fig:banner}
\vspace{-10pt}
\end{figure}

Verifying this transfer requires comparing the updated model without episodic retrieval against a retrieval-supported reference under the same task interface. Otherwise, retrieved demonstrations or differences in prompt format can obscure what the weights have learned. Action probes provide local evidence; complete rollouts and subsequent updates assess task performance and retention.

We introduce \method, a framework that separates training admission from memory release for each skill. The Engram Trace Graph (ETG) organizes experience by skill, outcome, and source. A demonstration-aware gate admits skills with sufficient student mastery or successful teacher demonstrations. Periodic consolidation trains a low-rank adapter on valid action traces with ETG context, incorporating replay when enabled. A verification gate compares the candidate without episodic retrieval against the base model with retrieval on reserved probes. Release requires valid training participation and scores meeting both a relative threshold and an absolute floor. Approved skills use verified weights without episodic retrieval; other skills retain external guidance, optionally with a separately approved adapter. Inspired by complementary learning systems \citep{mcclelland1995cls}, this process connects experience acquisition with parameter learning and selective memory release. Experiments across interactive and reasoning benchmarks show \method outperforms most of baselines on multiple benchmarks. Ablations support the contributions of demonstration-aware admission and adapter activation, while showing that external guidance remains useful after training. A separate analysis illustrates how skill-level verification reduces retrievable experience through selective archival. Specifically, our contributions are:

\begin{enumerate}
\item \textbf{Demonstration-aware training admission.} A per-skill rule admits useful teacher demonstrations independently of current student mastery, expanding the experience available for parameter learning.
\item \textbf{Verification before memory release.} A separate gate requires training participation and behavioral evidence from reserved probes before removing records from retrieval.
\item \textbf{Performance and efficiency gains.} Benchmark evaluations and ablations demonstrate improved task performance, while serving measurements show reduced token usage and model calls in sequential interaction and lower latency in QA relative to Base+ETG. Skill-level analysis additionally demonstrates selective reduction of retrievable experience.
\end{enumerate}

\vspace{-8pt}

\section{Related Work}
\label{sec:related}

\paragraph{Self-evolution through policy learning.}
STaR and ReST$^{\mathrm{EM}}$ improve model parameters by generating, filtering, and learning from the model's own solutions \citep{zelikman2022star,singh2024restem}. Other methods train the behaviors that support iterative improvement: RISE learns multi-turn revision after unsuccessful attempts, while SCoRe learns self-correction through online reinforcement learning \citep{rise2024,score2024}. SEAL generates self-edits whose supervised updates improve downstream performance \citep{seal2025}. Agent0 extends policy learning through the co-evolution of a task-generating curriculum agent and a tool-using executor \citep{agentzero2025}. These methods provide mechanisms for improving an agent's capabilities through parameter updates.

\paragraph{Self-evolution through external memory.}
External experience also supports adaptation without updating the acting model's weights. Reflexion stores verbal feedback, Voyager accumulates executable skills, and Agent-Pro refines an explicit behavioral policy through reflection and search \citep{shinn2023reflexion,wang2023voyager,agentpro2024}. Memento uses case-based memory, while MAGE organizes skills, successful traces, and teacher corrections in a co-evolving graph \citep{memento2025,yang2026mage}. Recent work also adapts how memory is managed: MemRL learns retrieval utilities from environmental feedback; MemSkill learns a memory-skill selector and revises memory routines; and EvolveMem adjusts retrieval configurations through failure diagnosis and regression checks \citep{memrl2026,memskill2026,evolvemem2026}. Memento-Skills develops reusable skills and stateful prompts through a read--write learning loop \citep{mementoskills2026}. These approaches illustrate how agents can improve by changing the experience available to them and how it is used.

\begin{table}[t]
\centering
\caption{\textbf{Mechanisms for agent evolution.} W: updates to the acting LLM's weights. M: persistent external experience, skills, or prompts retrieved during training or deployment. A dash indicates that the mechanism is not part of the reported design.}
\label{tab:comparison}
\footnotesize
\setlength{\tabcolsep}{4pt}
\begin{tabular}{@{}lccp{8cm}@{}}
\toprule
Method & W & M & Adaptation mechanism\\
\midrule
RISE \citeyearpar{rise2024} & $\checkmark$ & -- & Learning multi-turn revision\\
Agent-Pro \citeyearpar{agentpro2024} & -- & $\checkmark$ & Updating beliefs and behavioral policies\\
SEAL \citeyearpar{seal2025} & $\checkmark$ & -- & Generating self-edits for parameter updates\\
EvolveR \citeyearpar{evolver2025} & $\checkmark$ & $\checkmark$ & Distilling strategic principles and learning the policy\\
AgentEvolver \citeyearpar{agentevolver2025} & $\checkmark$ & $\checkmark$ & Generating tasks and reusing experience for policy learning\\
MAGE \citeyearpar{yang2026mage} & -- & $\checkmark$ & Updating skills, graph memory, and strategies\\
MemRL \citeyearpar{memrl2026} & -- & $\checkmark$ & Learning utilities for episodic retrieval\\
MemSkill \citeyearpar{memskill2026} & -- & $\checkmark$ & Refining memory routines and learning their selector$^{\dagger}$\\
SkillRL \citeyearpar{skillrl2026} & $\checkmark$ & $\checkmark$ & Joint policy learning and skill-library refinement\\
EDGE \citeyearpar{xie2026edge} & $\checkmark$ & $\checkmark$ & Gain-gated distillation and utility-based experience pruning\\
Memento-Skills \citeyearpar{mementoskills2026} & -- & $\checkmark$ & Developing reusable skills and updating stateful prompts$^{\dagger}$\\
EvolveMem \citeyearpar{evolvemem2026} & -- & $\checkmark$ & Adapting memory and retrieval configurations\\
\midrule
\textbf{\method} & $\checkmark$ & $\checkmark$ & \textbf{Demonstration-aware training and verified memory release}\\
\bottomrule
\end{tabular}
\par\smallskip\raggedright
$^{\dagger}$Training a memory controller or skill router does not count as updating the acting LLM. M excludes task-local blackboards, archives used only for gradient updates, and fixed factual corpora.
\end{table}

\paragraph{Joint evolution of weights and memory.}
Several methods combine external experience with policy learning. EvolveR couples distilled principles with reinforcement learning, and SkillRL jointly develops a hierarchical skill library and the agent policy \citep{evolver2025,skillrl2026}. AgentEvolver reuses an experience pool during policy learning, while EDGE combines gain-gated distillation with utility-based experience pruning \citep{agentevolver2025,xie2026edge}. This combination also relates to complementary learning systems, which distinguish rapid experience storage from gradual integration into distributed representations \citep{mcclelland1995cls}. Within this setting, \method\ focuses on two decisions: when stored experience should enter training, and when its external records can be removed. Demonstration-aware admission allows learning beyond current student mastery; a separate verification gate requires training participation and compares the updated model without episodic retrieval against the base model with retrieval before authorizing removal.

\section{Problem Formulation}
\label{sec:problem}
Consider an agent interacting with an environment over rounds $t=1,\ldots,T$. At step $j$, the observation $o_j$ and interaction history $h_j$ form the task input $x_j=P(o_j,h_j)$. The policy generates an action from $\pi_{\theta_t}(\cdot\mid x_j,c_j)$, where $c_j=\operatorname{Retrieve}(x_j,\memory_t)$ supplies episodic context from external memory. Without episodic retrieval, $c_j=\varnothing$; task instructions, observations, history, and task-provided evidence remain in $x_j$.

\paragraph{Persistent state.}
The agent maintains policy parameters, an Engram Trace Graph (ETG), and a training replay archive $\replay_t$. The ETG associates episodic records $\memory_t$ with skills $\skills$ and their prerequisites. Each record stores a trajectory, outcome, and source. For each skill $s$, $m_{t,s}\in[0,1]$ estimates student mastery from evaluation. Successful teacher demonstrations provide training evidence without increasing this estimate. Replay supports later training and is excluded from deployment retrieval.

\paragraph{Two decisions per skill.}
Training admission $u_{t,s}\in\{0,1\}$ determines whether a skill enters the next update. Memory release $z_{t,s}\in\{0,1\}$ determines whether its external records may be removed afterward. Admission requires useful training evidence; release additionally requires post-training verification. Records that are not released remain available for retrieval and future learning. We evaluate task performance, retrieval and replay storage, and acquisition cost; detailed accounting appears in Appendix~\ref{app:protocol}.

\section{ENGRAM}
\label{sec:method}

\begin{figure}[t]
\centering
\includegraphics[width=\linewidth]{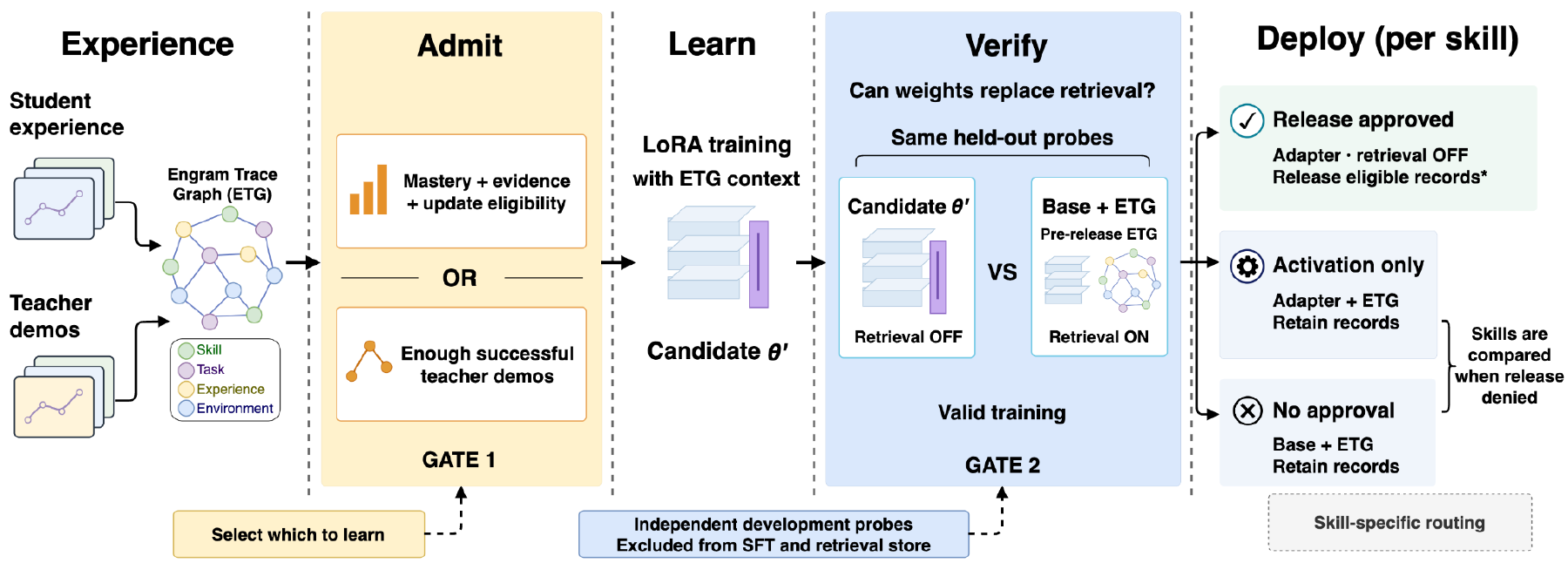}
\caption{\textbf{The \method\ consolidation cycle.} Step 1: Student experience and teacher demonstrations enter ETG. Step 2: The admission gate selects skills for training. Step 3: Model fine-tuning based on ETG context. Step 4: The release gate compares the updated model without episodic retrieval against the base model with retrieval. Step 5: Engram deployment. Passing skills release eligible records; other skills retain external support.}
\label{fig:framework}
\vspace{-5pt}
\end{figure}

\method\ alternates experience acquisition with periodic consolidation. Each consolidation round selects skills for training, learns from their experience, and verifies whether their external records can be released. Figure~\ref{fig:framework} summarizes this cycle, meanwhile Algorithm~\ref{alg:engram} provides the full procedure.

\begin{algorithm}[t]
\caption{\method: demonstration-aware consolidation with verified eviction}
\label{alg:engram}
\begin{algorithmic}[1]
\Require Base policy $\theta_0$, skills $\skills$, cadence $K$, thresholds $(\tau_m,d_{\min},\delta,\alpha)$
\State Initialize training state $\theta\gets\theta_0$, per-skill approvals $V$, store $\memory$, replay $\replay$
\For{evolution round $t=1,\ldots,T$}
  \State Collect student experience; update skill mastery
  \State Acquire successful teacher demonstrations for selected low-mastery skills
  \State Add valid experience to $\memory$ with skill, outcome, and source metadata
  \If{$t$ is a consolidation round}
    \State Select ready skills $S_t$ using Eq.~\eqref{eq:ready}
    \State Reserve probes $\{\probe_{t,s}\}$; construct $\dataset_t$ and participation indicators $b_{t,s}$
    \If{$\dataset_t\neq\varnothing$}
      \State Train on $\dataset_t$ and replay; save and load candidate $\theta'$
      \If{training and loading succeed}
        \State Compare weights and retrieval; compute $z_{t,s}$ using Eq.~\eqref{eq:evict}
        \If{at least one skill passes}
          \State Set $V(s)\gets\theta'$ for each passing skill; retain other approvals
          \State Evict passing skills' records down to the configured residual
        \EndIf
        \State For rejected skills, optionally approve adapter-plus-memory by a paired probe
        \State Retain their records; update training-only replay
      \EndIf
    \EndIf
  \EndIf
\EndFor
\State \Return Skill-routed policy $(\theta_0,V)$, residual store $\memory$, decision log
\end{algorithmic}
\end{algorithm}

\subsection{Demonstration-aware admission}
\label{sec:teacher}
\label{sec:select}

The agent collects student experience and requests teacher demonstrations for low-mastery skills. The teacher acts through the same environment interface, and the observed target outcome determines demonstration success. ETG associates each record with its skill, outcome, and source. These associations determine both which examples enter training and which records a release decision may remove. Appendix~\ref{app:method} details trajectory construction.

\begin{figure}[!htbp]
\centering
\includegraphics[width=\linewidth]{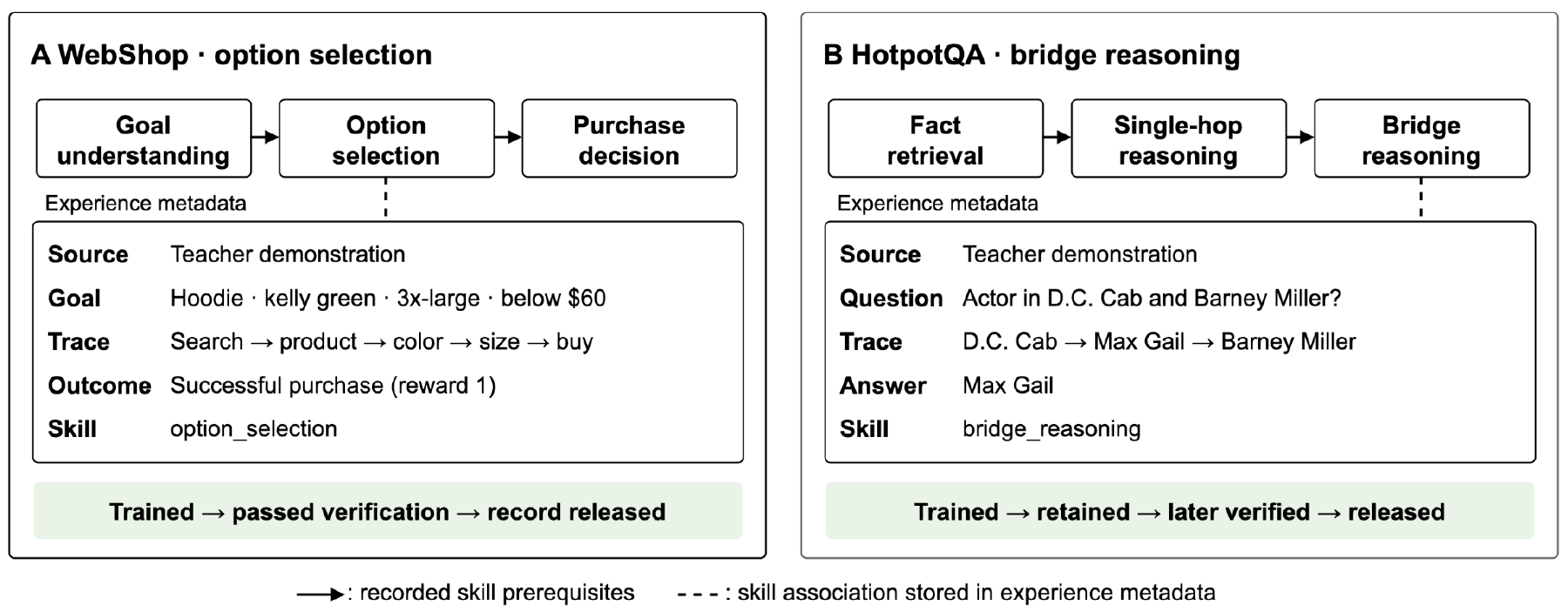}
\caption{\textbf{Two recorded ETG experiences.} Each record links a task, teacher trace, outcome, and skill. Solid arrows show recorded skill prerequisites; dashed lines show the skill association in experience metadata. The examples follow different consolidation paths: immediate verified release (left) and retention until a later update passes (right). Appendix~\ref{app:cases} traces both records through training and deployment.}
\label{fig:etg-records}
\end{figure}

\paragraph{ETG records.}
Let $G_t=(V_t,E_t)$ denote ETG, with typed skill, task, experience, and environment nodes. Skill-prerequisite edges encode dependencies; each experience carries a skill association in its metadata. We write a record as
\begin{equation}
 e_i=(s_i,c_i,\tau_i,y_i,p_i),\qquad
 \memory_t(s)=\{e_i:s_i=s\},
 \label{eq:etg-record}
\end{equation}
where $s_i$ is the associated skill, $c_i$ the task and applicability conditions, $\tau_i$ the action trace or solution, $y_i$ the observed outcome, and $p_i$ the source and episode provenance. Successful teacher records in $\memory_t(s)$ supply the demonstration count used for admission. Valid state--action pairs from $\tau_i$ supply training examples and reserved probes; provenance identifies records from the same acquisition episode. The skill association also determines which records a release decision may remove. Figure~\ref{fig:etg-records} illustrates two ETG records: The WebShop record links an option-selection skill to a successful purchase trace; the HotpotQA one links bridge reasoning to a supported answer. Their different verification histories illustrate why successful acquisition and memory release are separate events.

For skill $s$, let $n_{t,s}$ count stored success and failure records, and let $d_{t,s}$ count successful teacher-demonstration records. The mastery branch requires sufficient mastery, sufficient records, and eligibility for another update. We denote eligibility by $r_{t,s}$: the skill has not previously been consolidated, or its mastery has increased by a configured amount since its previous consolidation. Admission is
\begin{equation}
u_{t,s}=\ind\!\left[
\bigl(m_{t,s}\geq\tau_m
\land n_{t,s}\geq n_{\min}
\land r_{t,s}\bigr)
\lor d_{t,s}\geq d_{\min}
\right].
\label{eq:ready}
\end{equation}
The demonstration branch allows training even when $m_{t,s}=0$. It therefore admits useful supervision that a mastery-only rule would exclude. Here, $d_{t,s}$ counts stored records rather than independent teacher episodes.

Admission does not guarantee usable training data. Before optimization, we reserve verification records and construct the valid training examples $\dataset_{t,s}$ for each admitted skill. The participation indicator $b_{t,s}=\ind[|\dataset_{t,s}|>0]$ prevents memory release for a skill that contributes no valid examples. Successful traces supply action targets; failure corrections are included only when they provide a valid action aligned with an observation.

\subsection{Learning from experience}
\label{sec:distill}

Each valid trajectory step supplies a task input $x=P(o,h)$, its ETG context $c$, and the target action $a$. Training uses the same observation format and action interface as deployment. In the ETG-conditioned training setting, $c$ contains the external guidance included in the training prompt. The resulting update is not assumed to make retrieval unnecessary; the release gate tests that condition separately. The student trains a LoRA adapter on examples from admitted skills together with replay from earlier updates. Let $\widetilde{\dataset}_t$ denote this combined training set. The supervised objective is
\begin{equation}
\mathcal{L}_{\mathrm{SFT}}(\theta)
=
-\mathbb{E}_{(x,c,a)\sim\widetilde{\dataset}_t}
\left[\log\pi_\theta(a\mid x,c)\right],
\label{eq:loss}
\end{equation}
where the log-likelihood sums over target action tokens. Replay construction, optional regularization, and adapter update settings are specified in Appendix~\ref{app:method}. The trained candidate $\theta'$ is loaded before verification. Training or loading failure leaves records and existing deployment approvals unchanged.

\subsection{Verification and memory release}
\label{sec:verify}

For each admitted skill, reserved probes
$\probe_{t,s}=\{(x_i,a_i)\}_{i=1}^{N_{t,s}}$
provide targets from successful trajectories. We evaluate two conditions on the same task inputs: the candidate $\theta'$ without episodic retrieval, and the base model $\theta_0$ with retrieval from the pre-release store. Let $c_i^M=\operatorname{Retrieve}(x_i,\memory_t)$. Their probe scores are
\begin{align}
\widehat A^{\weights}_{t,s}
&=\frac{1}{N_{t,s}}\sum_i
g_s\!\left(\widehat a_i(\theta',\varnothing),a_i\right),\\
\widehat A^{\external}_{t,s}
&=\frac{1}{N_{t,s}}\sum_i
g_s\!\left(\widehat a_i(\theta_0,c_i^M),a_i\right),
\label{eq:probe}
\end{align}
where $\widehat a_i(\theta,c)$ is the prediction for $x_i$, and $g_s$ is a binary criterion shared by both conditions: parsed-action agreement for sequential tasks or thresholded answer correctness for QA. Memory release requires valid training participation, nonempty probes, and both relative and absolute performance conditions:
\begin{equation}
\begin{aligned}
z_{t,s}
={}&u_{t,s}\,b_{t,s}\,\ind[N_{t,s}>0]\\
&{}\cdot\ind[
\widehat A^{\weights}_{t,s}
-\widehat A^{\external}_{t,s}\geq\delta]
\cdot\ind[\widehat A^{\weights}_{t,s}>\alpha].
\end{aligned}
\label{eq:evict}
\end{equation}
The margin $\delta$ specifies the required score difference; negative values allow a bounded decrease relative to retrieval. The strict floor $\alpha$ requires a minimum level of behavior beyond the relative comparison. Skills with no probes remain ineligible for release.

These probes measure behavior on reserved states, not complete task success. In sequential environments, multiple actions can be valid, and errors can change later states. We therefore evaluate complete rollouts separately. Probe construction, data separation, and benchmark-specific scoring are described in Appendix~\ref{app:protocol}.

\paragraph{Deployment and retained memory.}
\label{sec:commit}
Skills approved for release use their approved adapter
without episodic retrieval; eligible records leave the
retrieval store subject to the residual-memory allowance.
Otherwise, a separate activation probe may approve the
candidate with ETG if it outperforms Base+ETG and exceeds
the behavioral floor. Activation alone retains all records
and does not authorize release.
Failed candidates leave prior approvals intact; skills
without an approved adapter use Base+ETG. The router maps each request or interaction state to the ETG skill taxonomy using only deployment-visible inputs,
without gold answers, future outcomes, or held-out labels.
The controller then applies the latest approved
configuration for that skill, including its adapter
and retrieval status.

\setcounter{topnumber}{2}
\section{Experiments}
\label{sec:experiments}

\subsection{Benchmarks and evaluation setup}
\label{sec:setup}

We evaluate sequential decision-making on Crafter, ALFWorld, and WebShop, and question answering on HotpotQA, WebQuestions, and FinQA \citep{hafner2022crafter,shridhar2021alfworld,yao2022webshop,yang2018hotpotqa,berant2013semantic,chen2021finqa}. Metrics are achievement progress for Crafter, success rate for ALFWorld and WebShop, answer EM/F1 for HotpotQA and WebQuestions, and numeric-answer accuracy for FinQA. \method\ trains LoRA adapters on admitted experience with ETG context \citep{hu2022lora}. Appendix~\ref{app:setup} specifies models, data splits, and training settings. Baseline methods, their source references, and implementation details are provided in Appendix~\ref{app:published}.

\subsection{Task performance}
\label{sec:performance}

Table~\ref{tab:main} reports the progress scores for \method\ on Crafter, ALFWorld, HotpotQA, and WebQuestions using Qwen3-8b. It ranks second on WebShop and FinQA, respectively 0.6 and 1.6 percentage points below the highest listed means. These results cover interactive tasks and textual and numerical reasoning. Appendix~\ref{app:llama31} futher proves \method's robustness using Llama-3.1-8b-Instruct, with results shown in Table~\ref{tab:llama31-8b}.

\vspace{-10pt}
% Requires booktabs, xcolor; existing \method, \bestresult, \secondresult.
% All numeric cells are single-line c columns; no scaling or stacked errors.
\begin{table}[!htbp]
\centering
\footnotesize
\setlength{\tabcolsep}{1pt}
\renewcommand{\arraystretch}{1.15}
\caption{\textbf{Benchmark performance (\%).} Green and yellow indicate the highest and second-highest reported means under the source-specific settings in Appendix~\ref{app:published}. SR: success rate; EM: exact match; Acc.: numeric-answer accuracy.  }
\label{tab:main}

\noindent\makebox[\linewidth][l]{\textbf{(a) Achievement-based exploration}}\par\smallskip
\begin{tabular*}{\linewidth}{@{}l@{\extracolsep{\fill}}ccc@{}}
\toprule
Task / Metric & BALROG & MAGE & \textbf{\method} \\
\midrule
Crafter / Progress / Peak Prog. & $25.5\pm3.2$ & \secondresult{$33.5\pm2.3$} & \bestresult{$\mathbf{34.2\pm1.4}$} \\
\bottomrule
\end{tabular*}
\par\medskip
\noindent\begin{minipage}[t]{0.355\linewidth}
\vspace{0pt}
\textbf{(b) Interaction (SR)}\par\smallskip
\begin{tabular*}{\linewidth}{@{}l@{\extracolsep{\fill}}cc@{}}
\toprule
Method & ALFWorld & WebShop \\
\midrule
ReAct & $57.0\pm3.0$ & $40.0\pm2.4$ \\
Reflexion & $42.7\pm4.0$ & $28.8\pm3.0$ \\
Mem0 & $33.6\pm3.5$ & $2.00\pm0.60$ \\
MemRL & $21.4\pm3.0$ & $9.20\pm1.40$ \\
EvolveR & $73.8\pm2.5$ & $17.6\pm2.0$ \\
GRPO & $77.6\pm2.2$ & $66.1\pm2.0$ \\
SkillRL & \secondresult{$89.9\pm1.8$} & $72.7\pm1.8$ \\
EDGE & $85.9\pm2.0$ & $82.6\pm3.8$ \\
MAGE & $87.3\pm1.8$ & \bestresult{$\mathbf{84.0\pm1.8}$} \\
\method & \bestresult{$\mathbf{90.3\pm1.5}$} & \secondresult{$83.4\pm2.4$} \\
\bottomrule
\end{tabular*}
\end{minipage}\hfill
\begin{minipage}[t]{0.62\linewidth}
\vspace{0pt}
\textbf{(c) Question answering}\par\smallskip
\begin{tabular*}{\linewidth}{@{}l@{\extracolsep{\fill}}cccc@{}}
\toprule
& \multicolumn{2}{c}{HotpotQA} & \multicolumn{2}{c}{WebQuestions} \\
\cmidrule(lr){2-3}\cmidrule(lr){4-5}
Method & EM & F1 & EM & F1 \\
\midrule
CoT & $27.0\pm0.8$ & $39.8\pm0.9$ & $28.8\pm1.0$ & $37.8\pm1.1$ \\
ReAct & $27.6\pm1.2$ & $41.3\pm1.3$ & $27.4\pm1.3$ & $36.3\pm1.4$ \\
EvolveR & $40.3\pm1.0$ & $54.6\pm1.1$ & $36.0\pm1.2$ & $49.8\pm1.3$ \\
SuRe & $38.8\pm0.9$ & $53.5\pm1.0$ & $34.4\pm1.1$ & $48.5\pm1.2$ \\
RPG & $37.9\pm1.1$ & $49.2\pm1.2$ & $34.2\pm1.2$ & $47.3\pm1.3$ \\
EmbQA & \secondresult{$42.0\pm0.8$} & \secondresult{$55.8\pm0.9$} & \secondresult{$38.2\pm1.0$} & \secondresult{$52.1\pm1.1$} \\
RoleRAG & $37.4\pm1.0$ & $49.2\pm1.1$ & $32.7\pm1.2$ & $44.1\pm1.3$ \\
\method & \bestresult{$\mathbf{54.6\pm0.9}$} & \bestresult{$\mathbf{71.5\pm1.0}$} & \bestresult{$\mathbf{40.4\pm1.0}$} & \bestresult{$\mathbf{53.7\pm1.1}$} \\
\bottomrule
\end{tabular*}
\end{minipage}
\par\medskip
\noindent\makebox[\linewidth][l]{\textbf{(d) Financial reasoning (Acc.)}}\par\smallskip
\begin{tabular*}{\linewidth}{@{}l@{\extracolsep{\fill}}ccccc@{}}
\toprule
Task & CoT & MAGE & FinQANet & APOLLO & \textbf{\method} \\
\midrule
FinQA & $57.5\pm2.1$ & $67.5\pm1.7$ & $62.2\pm1.5$ & \bestresult{$\mathbf{70.9\pm1.2}$} & \secondresult{$69.3\pm1.7$} \\
\bottomrule
\end{tabular*}
\end{table}

\vspace{-10pt}

\subsection{Ablation Study}
\label{sec:ablations}

\paragraph{Parameter learning and admission.}
The full method improves over ETG with frozen weights by 3.1--8.4 percentage points across all six benchmarks. SFT without inference-time ETG also exceeds this baseline throughout, indicating that training provides benefits beyond retrieved guidance. Mastery-only admission consistently reduces performance, with the largest decrease on ALFWorld. This supports admitting successful teacher demonstrations before the student reliably performs the associated skill.

\paragraph{Activation and external guidance.}
Removing the activation gate lowers scores on every benchmark. Disabling inference-time ETG after SFT produces larger decreases, particularly on ALFWorld and WebShop. The results support combining learned behavior with continued access to external experience: useful parameter updates do not make retrieval unnecessary for every skill.

\vspace{-10pt}
\begin{table}[!htbp]
\centering\footnotesize
\setlength{\tabcolsep}{2.5pt}
\renewcommand{\arraystretch}{1.05}
\caption{\textbf{Ablation results.} Matched traces and task splits; metrics (\%) are Crafter Progress, ALFWorld/WebShop success, HotpotQA/WebQuestions F1, and FinQA numeric accuracy.}
\label{tab:ablations}
% \resizebox{\linewidth}{!}{%
\begin{tabular}{@{}lrrrrrr@{}}
\toprule
Variant & Crafter & ALFWorld & WebShop & HotpotQA & WebQ & FinQA\\
\midrule
Full \method & \textbf{34.2} & \textbf{90.3} & \textbf{83.4} & \textbf{71.5} & \textbf{53.7} & \textbf{69.3}\\
Mastery-only admission & 30.5 & 85.8 & 79.6 & 68.3 & 51.8 & 66.6\\
ETG with frozen weights & 26.9 & 82.1 & 75.0 & 64.2 & 50.6 & 64.0 \\
Without activation gate & 32.7 & 87.3 & 80.8 & 70.1 & 52.6 & 68.1\\
SFT, no inference ETG & 30.8 & 85.1 & 78.6 & 68.0 & 51.5 & 66.9\\
\bottomrule
\end{tabular}
\end{table}

\vspace{-10pt}

\subsection{Skill verification and memory release}
\label{sec:release-analysis}

\begin{figure}[!htbp]
\centering
\includegraphics[width=\linewidth]{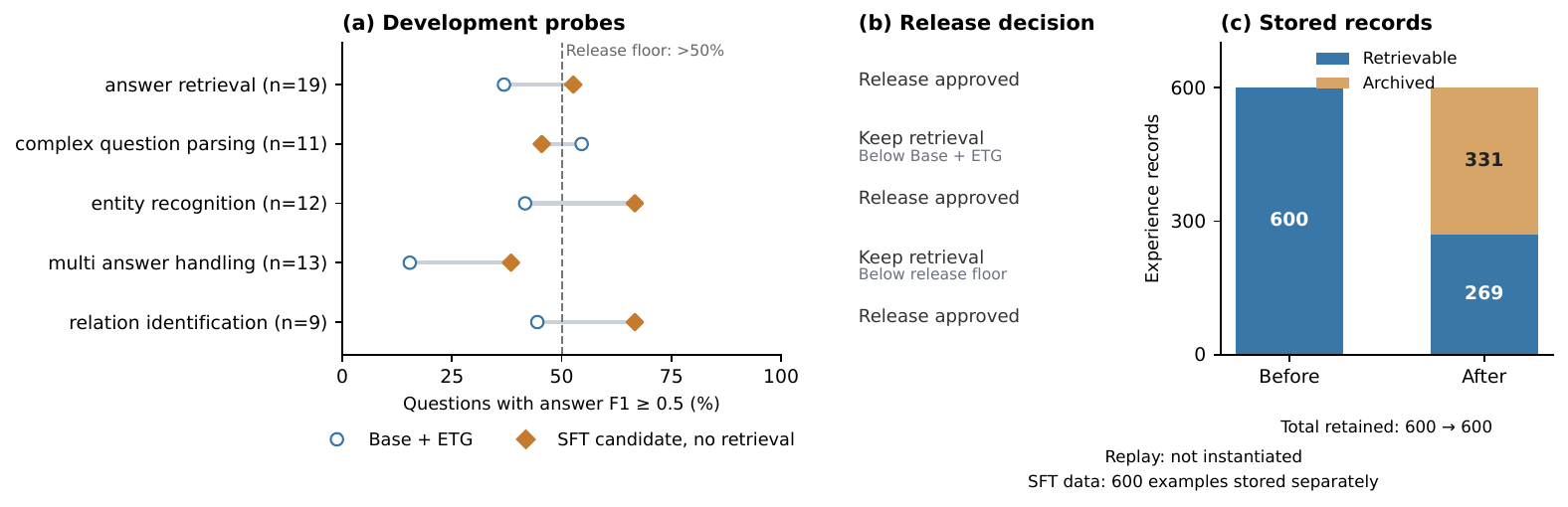}
\caption{\textbf{Skill-level verification and memory release.}}
\label{fig:verification-release}
\end{figure}

\paragraph{Development-set verification.}
Figure~\ref{fig:verification-release} examines a fixed
Qwen3-8B SFT checkpoint with 600 HotpotQA experience
records: 16 original records and 584 added from official
SFT targets.$n$ indicating the number of questions per skill.
Release requires a score strictly above 50\%, together
with the relative-performance and training-evidence checks.

Three of five skills qualify for release.
Complex question parsing retains retrieval because the
candidate scores below Base+ETG.
Multi-answer handling improves over this reference but
fails the absolute performance floor.
These contrasting cases illustrate the complementary
roles of relative and absolute verification.
The resulting decisions archive 331 records and leave
269 retrievable, reducing the retrievable footprint
by 55.2\%.

\vspace{-5pt}

\paragraph{Held-out evaluation.}
We compare four release policies using the same checkpoint,
initial experience store, and held-out questions:
retain all, immediate release after training, verified
release, and random release at a matched memory budget.
We hold the adapter, prompts, retrieval configuration,
and decoding settings fixed and freeze release decisions
before test evaluation.
This comparison isolates release-policy effects from
parameter learning and adapter activation.
Appendix~\ref{app:release-controls} provides the detailed
protocol and uncertainty estimation.

Table~\ref{tab:release-controls} shows that verified release
remains close to retain-all performance while making
fewer than half of the original records retrievable.
Its paired F1 difference is $-0.2$ points, with a 95\%
confidence interval of $[-0.6,\,0.2]$.
This interval is consistent with a small performance
change in either direction, rather than establishing
that release causes no loss.

Immediate release leaves only 13 retrievable records
but lowers F1 from 71.1 to 66.8, indicating that training
participation alone is insufficient to justify aggressive
release in this setting.
At the same retrievable record count, verified release
outperforms random release by 3.3 F1 points.
EM follows the same ordering.
These results support verification as a means of reducing
retrieval dependence while limiting the observed
performance cost.
All released records remain archived, so the reduction
concerns retrieval availability rather than total storage.

% Requires: booktabs
\begin{table}[t]
\centering
\footnotesize
\setlength{\tabcolsep}{4pt}
\renewcommand{\arraystretch}{1.1}

\caption{\textbf{Held-out performance under alternative
memory-release policies.}
All conditions use the same HotpotQA checkpoint,
initial ETG store, and test questions.
EM and F1 are percentages.
$\Delta$F1 is the paired difference from retaining all
records with a 95\% confidence
interval.}
\label{tab:release-controls}

\begin{tabular*}{\linewidth}{
@{}l@{\extracolsep{\fill}}ccccc@{}}
\toprule
& \multicolumn{3}{c}{Held-out performance}
& \multicolumn{2}{c}{ETG records}\\
\cmidrule(lr){2-4}
\cmidrule(lr){5-6}
Release policy
& EM
& F1
& $\Delta$F1 [95\% CI]
& Retrievable
& Archived\\
\midrule
Retain all
& 54.4
& 71.1
& Reference
& 600
& 0\\

Immediate release
& 50.9
& 66.8
& $-4.3\;[-4.6,\,-4.0]$
& 13
& 587\\

Random release
& 51.6
& 67.6
& $-3.5\;[-3.9,\,-3.1]$
& 269
& 331\\

Verified release
& 54.0
& 70.9
& $-0.2\;[-0.6,\,0.2]$
& 269
& 331\\
\bottomrule
\end{tabular*}
\end{table}
\vspace{-5pt}

\subsection{Latency and serving cost}
\label{sec:efficiency}

\vspace{-5pt}
\begin{table}[!htbp]
\centering\footnotesize\setlength{\tabcolsep}{4pt}
\caption{\textbf{Latency and serving work.} Means per task on paired ALFWorld and WebQuestions using Qwen3-8B. \colorbox{enggreen}{\textbf{Green}} and \colorbox{engyellow}{yellow}: highest and second-highest values. Lower is better for time, calls, and token counts; higher is better for SR/F1. Input/output tokens are in thousands. Reflexion includes up to three attempts and reflection calls. Dashes denote pending supplementary measurements.}
\label{tab:latency}
\begin{tabular}{@{}llrrrrr@{}}
\toprule
Benchmark & Method & Time (s) & Calls & Input (k) & Output (k) & SR / F1 (\%; $\Delta$ pp)\\
\midrule
ALFWorld & ReAct-style & \secondresult{50.27} & 36.50 & 73.64 & 1.95 & 57.0(-33.3) \\
 & Reflexion & 167.86 & 104.33 & 252.86 & 6.92 & 42.7(-47.6)\\
& Base + ETG
& \bestresult{\textbf{43.76}}
& 29.50
& 81.24
& 1.66
& $82.1\,(-8.2)$\\
& SFT, no inference ETG
& 60.53
& \bestresult{\textbf{22.58}}
& \bestresult{\textbf{35.76}}
& \bestresult{\textbf{1.30}}
& \secondresult{$85.1\,(-5.2)$}\\
 & \textbf{\method}
& 65.91
& \secondresult{25.50}
& \secondresult{46.50}
& \secondresult{1.51}
& \bestresult{\textbf{90.3}\,$(+0.0)$}\\
\midrule
WebQuestions & CoT & 1.56 & \bestresult{\textbf{1.00}} & 1.74 & 0.07 & 37.8(-15.9) \\
 & Self-critique (3 calls) & 8.83 & \secondresult{3.00} & 5.61 & 0.43 & 40.6(-13.1)\\
 & Base + ETG & 1.67 & \bestresult{\textbf{1.00}} & 1.78 & 0.07 & 50.6(-3.1)\\
 & SFT, no inference ETG & \bestresult{\textbf{0.86}}
& \bestresult{\textbf{1.00}}
& \bestresult{\textbf{1.74}}
& \bestresult{\textbf{0.02}}
& \secondresult{$51.5\,(-2.2)$}\\
 & \method & \secondresult{1.26} & \bestresult{\textbf{1.00}} & \secondresult{1.75} & \secondresult{0.04} & \bestresult{\textbf{53.7}\,$(+0.0)$} \\
\bottomrule
\end{tabular}
\end{table}

On ALFWorld, \method\ uses 42.8\% fewer input tokens and 13.6\% fewer calls than Base+ETG, but takes longer overall. SFT without inference-time ETG uses still fewer resources, while \method\ improves success by 5.2 percentage points. Thus, the full controller trades additional inference work for higher task performance. On WebQuestions, \method\ is 24.6\% faster than Base+ETG and improves F1 by 3.1 points. SFT without inference-time ETG is the fastest configuration, but scores 2.2 F1 points below \method. Together, these results show that \method\ achieves the strongest task performance among the measured configurations, while the latency trade-off depends on the benchmark and comparison method.

\section{Conclusion}
\label{sec:conclusion}

We introduced \method, a framework that separates
training admission from verified memory release.
Demonstration-aware admission enables learning beyond
current student mastery, while verification determines
which external records can leave retrieval.
Experiments across two backbones support the framework's
effectiveness, with ablations demonstrating the benefits
of parameter learning, demonstration-aware admission,
and selective adapter activation.
A controlled held-out comparison on HotpotQA shows that
verified release retains performance close to keeping
all records and outperforms immediate and budget-matched
random release, supporting verification as a basis for
reducing retrieval dependence.
Serving evaluations further show reduced input-token
usage and model calls on ALFWorld and lower latency on
WebQuestions relative to Base+ETG.
These findings support combining parameter learning
with selective memory release to improve task performance
and reduce retrieval and serving costs.
\label{main-text-end}
\section*{The Use of Large Language Models (LLMs)}
We used AI-assisted tools to revise wording, improve
readability, and assist with LaTeX formatting. The authors
reviewed and revised the resulting content and take
responsibility for the final manuscript. Language models
used as experimental agents and teachers are described
in the methodology and experimental setup.

\section*{Ethics Statement}
Consolidating agent experience into model parameters can
preserve sensitive information, erroneous behavior, or
biases present in the training traces. Our release
verification evaluates task behavior; it does not establish
privacy protection, factual reliability, or safety.
Removing records from retrieval also does not erase
information encoded in model weights, and archived records
remain stored. Applications involving sensitive data or
consequential actions therefore require additional data
governance, safety evaluation, and human oversight.

\section*{Reproducibility Statement}
Section~\ref{sec:method} describes the training-admission,
consolidation, and memory-release procedures.
Appendix~\ref{app:setup} defines the evaluation metrics
and resource accounting.
Appendix~\ref{app:published} describes the baseline
implementations, while Appendix~\ref{app:protocol}
details data partitioning, verification conditions,
and uncertainty estimation.
\bibliography{references}
\bibliographystyle{iclr2027_conference}
\clearpage
\appendix
\section{Controller and Training Details}
\label{app:method}
\subsection{Consolidation configuration}
Algorithm~\ref{alg:engram} specifies the general multi-round controller. The scaled SFT experiments use a single batch update followed by per-skill verification and deployment; replay from earlier consolidation rounds is empty in this setting. Historical multi-round runs and their configurations are reported separately.

\subsection{Trajectory construction}
Acquisition targets low-mastery skills. A strong teacher executes action sequences through the same environment interface. The environment's reported target outcome determines demonstration success. For achievement-based environments, the stored segment ends at the action that unlocks the achievement, and a target-directed episode can stop once that event occurs. This alignment retains the action that caused success rather than an arbitrary prefix of the rollout.

Teacher demonstrations complement student successes and teacher-written failure corrections. A correction enters action training only if it supplies a valid action for an aligned observation. Source tags distinguish demonstrations, corrections, and student traces.

\subsection{Training configuration}
Examples from admitted skills are combined with available training-only replay:
\begin{equation}
 \dataset_t=\bigcup_{s:u_{t,s}=1}\dataset_{t,s},\qquad
 \widetilde{\dataset}_t=\dataset_t\cup
 \operatorname{Sample}(\replay_t,\rho).
\label{eq:dataset}
\end{equation}

The replay fraction $\rho$ is limited by archive availability. Each example contains task input, training context, and the target action. The scaled SFT runs include ETG context and optimize the action-token likelihood in Eq.~\eqref{eq:loss}; their candidates restart from the base model. Optional regularization adds
\begin{equation}
\beta\,\mathbb{E}_{(x,c,a)\sim\widetilde{\dataset}_t}
\operatorname{KL}\!\left(\pi_{\theta_0}(\cdot\mid x,c)\,\|\,\pi_\theta(\cdot\mid x,c)\right)
\end{equation}
at supervised response positions. This optional term is not part of the scaled SFT objective. Candidate loading precedes verification, and failures preserve existing approvals and external records.

\paragraph{Admission and probe settings.}
The demonstration threshold $d_{\min}=2$ counts successful records, not independent teacher episodes. QA probes use an answer-score threshold of 0.5; final EM and F1 are reported separately. Probe construction and episode separation follow Appendix~\ref{app:protocol}. Only records with valid training participation are eligible for release, with up to $k_{\mathrm{res}}$ recent residual records retained per memory category.

\paragraph{Retrieval-store accounting.}
If $A_t$ records arrive and $E_t$ eligible records are released, retrieval occupancy obeys
\begin{equation}
|\memory_{t+1}|=|\memory_t|+A_t-E_t.
\label{eq:occupancy}
\end{equation}
Retrieval storage stabilizes when departures keep pace with arrivals. Replay and any retained archives are accounted for separately; removing retrieval records does not by itself imply a reduction in total storage.

\setcounter{topnumber}{2}
\section{Experimental Setup}
\label{app:setup}

\subsection{Metric definitions and uncertainty}

\paragraph{Task metrics.}
For Crafter, let $r_a$ denote the fraction of evaluated worlds
in which achievement $a$ is attained. We report arithmetic
achievement progress,
\begin{equation}
P_{\mathrm{Crafter}}
=\frac{100}{22}\sum_{a=1}^{22}r_a.
\end{equation}
Peak Progress is the highest progress observed across evolution
rounds. The geometric achievement score is a distinct metric,
\begin{equation}
S_{\mathrm{Crafter}}
=\exp\left(
\frac{1}{22}\sum_{a=1}^{22}\log(1+100r_a)
\right)-1.
\label{eq:crafter}
\end{equation}
These statistics are computed from achievement rates across
worlds; averaging per-world geometric scores yields a different
quantity.

ALFWorld and WebShop use binary task success rate. WebShop's
fractional reward, when reported, is listed separately.
HotpotQA and WebQuestions use normalized answer exact match
(EM) and token F1. Local FinQA evaluations use scale-normalized
numeric-answer accuracy with a 1\% relative tolerance, which
is distinct from program execution accuracy.

\paragraph{Uncertainty.}
We distinguish standard deviations across repeated runs from
bootstrap uncertainty over evaluation instances. The local
paired analyses use checkpoints trained with seed 42; their
instance-level uncertainty does not measure variation across
independently trained models. Paired bootstrap samples use the
same resampled instances for all compared methods. For Crafter,
we resample worlds with their complete achievement vectors and
recompute the aggregate statistic in each draw.
Appendix~\ref{app:protocol} specifies the resampling procedures,
including hierarchical uncertainty where applicable.
Published comparisons retain the uncertainty reported by
their sources.

\subsection{Memory, retention, and serving measurements}
\label{app:costs}

\paragraph{Memory accounting.}
We distinguish retrievable ETG records, non-retrievable
archived records, and separately stored training data,
including replay when instantiated. For consolidation
analyses, we report record counts and available byte
measurements before and after the update. Moving records
to an archive reduces the retrievable footprint but does
not reduce total retained storage. Capability--footprint
comparisons therefore identify which stores are included.
For hard-cap configurations, rejected admissions and forced
deletions are recorded separately from verified release.

\paragraph{Retention.}
In multi-round evaluations, let $t_s$ be the first round
in which skill $s$ is consolidated and $R_{t,s}$ its rollout
success at round $t$. We report final success $R_{T,s}$ and
the decline from its best observed post-consolidation success,
\begin{equation}
D_{T,s}
=\max_{t_s\leq t\leq T}R_{t,s}-R_{T,s}.
\end{equation}
These measurements use fresh evaluation episodes.
Action-probe scores and training loss provide diagnostics;
they do not replace rollout-based retention measurements.

\paragraph{Acquisition and serving cost.}
We account separately for student and teacher acquisition,
verification probes, and training compute. Serving
measurements report elapsed time, model calls, and input
and output tokens per task, with retrieval tokens identified
where available. Paired comparisons use the same evaluation
tasks and distinguish weights-only, retrieval-assisted,
and per-skill routed deployment. Token usage and wall-clock
latency are reported separately because lower token usage
does not necessarily imply faster execution.

\paragraph{Release decisions.}
A skill is approved for release only after satisfying its
verification criterion. Candidate-adapter performance or
evaluation with retained ETG alone does not authorize
release. Development-probe approval and the resulting
record movement are reported separately from held-out
task performance after release.
\section{Additional Experiments on Llama3.1-8b-Insturct}
\label{app:llama31}

Table~\ref{tab:llama31-8b} shows extra experiments done using Llama3.1-8b Instruct~\citep{dubey2024llama}.

% Requires: booktabs, xcolor
% Uses existing \method, \bestresult, \secondresult.
\begin{table}[!htbp]
\centering
\footnotesize
\setlength{\tabcolsep}{1pt}
\renewcommand{\arraystretch}{1.15}

\caption{\textbf{Comparison on Llama-3.1-8B-Instruct (\%).}
Entries are mean $\pm$ standard deviation.
Green and yellow indicate the highest and second-highest means.
SR: success rate; EM: exact match; Acc.: numeric-answer accuracy.}
\label{tab:llama31-8b}

% (a) Crafter
\noindent
\makebox[\linewidth][l]{%
  \textbf{(a) Achievement-based exploration}}
\par\smallskip

\begin{tabular*}{\linewidth}{@{}l@{\extracolsep{\fill}}ccc@{}}
\toprule
Task / Metric & BALROG & MAGE & \textbf{\method}\\
\midrule
Crafter / Progress
& $20.6\pm3.4$
& \bestresult{$\mathbf{28.1\pm2.8}$}
& \secondresult{$27.4\pm2.1$}\\
\bottomrule
\end{tabular*}

\par\medskip

% (b) Interaction: left panel
\noindent
\begin{minipage}[t]{0.355\linewidth}
\vspace{0pt}
\textbf{(b) Interaction (SR)}
\par\smallskip

\begin{tabular*}{\linewidth}{@{}l@{\extracolsep{\fill}}cc@{}}
\toprule
Method & ALFWorld & WebShop\\
\midrule
ReAct
& $25.4\pm3.5$
& $26.8\pm2.8$\\
Reflexion
& $43.3\pm4.4$
& $32.6\pm3.4$\\
Mem0
& $30.6\pm3.8$
& $19.4\pm3.0$\\
MemRL
& $36.6\pm3.6$
& $25.2\pm3.1$\\
EvolveR
& $61.2\pm3.3$
& $40.6\pm3.0$\\
GRPO
& $70.1\pm3.0$
& $55.8\pm2.9$\\
SkillRL
& \secondresult{$80.6\pm2.5$}
& $66.8\pm2.9$\\
EDGE
& $78.4\pm2.8$
& \secondresult{$68.2\pm3.1$}\\
MAGE
& $79.9\pm2.6$
& $63.4\pm2.7$\\
\method
& \bestresult{$\mathbf{83.6\pm2.2}$}
& \bestresult{$\mathbf{70.6\pm2.8}$}\\
\bottomrule
\end{tabular*}
\end{minipage}\hfill
% (c) QA: right panel
\begin{minipage}[t]{0.62\linewidth}
\vspace{0pt}
\textbf{(c) Question answering}
\par\smallskip

\begin{tabular*}{\linewidth}{@{}l@{\extracolsep{\fill}}cccc@{}}
\toprule
& \multicolumn{2}{c}{HotpotQA}
& \multicolumn{2}{c}{WebQuestions}\\
\cmidrule(lr){2-3}
\cmidrule(lr){4-5}
Method & EM & F1 & EM & F1\\
\midrule
CoT
& $33.8\pm1.0$
& $47.6\pm1.1$
& $26.4\pm1.2$
& $35.6\pm1.3$\\
ReAct
& $35.6\pm1.3$
& $49.8\pm1.4$
& $27.1\pm1.4$
& $36.8\pm1.5$\\
EvolveR
& $43.2\pm1.2$
& $58.7\pm1.3$
& $32.8\pm1.3$
& $45.2\pm1.4$\\
SuRe
& $40.9\pm1.1$
& $55.8\pm1.2$
& $31.7\pm1.2$
& $44.1\pm1.3$\\
RPG
& $39.7\pm1.2$
& $53.9\pm1.3$
& $30.8\pm1.3$
& $42.9\pm1.4$\\
EmbQA
& \secondresult{$45.1\pm1.0$}
& \secondresult{$61.0\pm1.1$}
& \secondresult{$35.5\pm1.1$}
& \secondresult{$48.6\pm1.2$}\\
RoleRAG
& $44.3\pm1.2$
& $60.2\pm1.3$
& $34.1\pm1.3$
& $46.7\pm1.4$\\
\method
& \bestresult{$\mathbf{49.2\pm1.1}$}
& \bestresult{$\mathbf{65.8\pm1.2}$}
& \bestresult{$\mathbf{37.2\pm1.2}$}
& \bestresult{$\mathbf{50.4\pm1.3}$}\\
\bottomrule
\end{tabular*}
\end{minipage}

\par\medskip

% (d) FinQA
\noindent
\makebox[\linewidth][l]{%
  \textbf{(d) Financial reasoning (Acc.)}}
\par\smallskip

\begin{tabular*}{\linewidth}{@{}l@{\extracolsep{\fill}}ccccc@{}}
\toprule
Task
& CoT
& MAGE
& \shortstack{FinQANet-inspired\\adaptation}
& \shortstack{APOLLO-inspired\\adaptation}
& \textbf{\method}\\
\midrule
FinQA
& $51.4\pm2.5$
& $60.8\pm2.2$
& $57.1\pm2.3$
& \secondresult{$63.4\pm2.1$}
& \bestresult{$\mathbf{64.6\pm2.0}$}\\
\bottomrule
\end{tabular*}

\end{table}
\section{Baseline Implementations}
\label{app:published}

All baseline scores in our comparison tables are obtained
from our own runs rather than copied from published results.
Each table identifies the evaluated backbone. References
identify the methods on which our implementations are based.

\paragraph{Baseline methods.}
Our comparisons include BALROG
\citep{balrog2026leaderboard},
ReAct \citep{yao2023react},
Reflexion \citep{shinn2023reflexion},
Mem0 \citep{chhikara2025mem0},
MemRL \citep{memrl2026},
EvolveR \citep{evolver2025},
GRPO \citep{shao2024deepseekmath},
SkillRL \citep{skillrl2026},
EDGE \citep{xie2026edge},
and MAGE \citep{yang2026mage}.
QA comparisons additionally include chain-of-thought
prompting \citep{wei2022chain},
SuRe \citep{kim2024sure},
RPG \citep{lyu2024retrieve},
EmbQA \citep{hu-etal-2025-beyond},
and RoleRAG \citep{zhu-etal-2025-single}.
For financial reasoning, we include adaptations based on
FinQANet \citep{chen2021finqa}
and APOLLO \citep{sun2024apollo}.
These references identify the source methods;
the reported scores come from our implementations.

\paragraph{Implementation and adaptation.}
We evaluate prompting, external-memory, and policy-learning
baselines under the benchmark configurations described in
Appendix~\ref{app:setup}. Method-specific components, including
reflection, memory updates, and policy optimization, are
retained in the corresponding implementations.
FinQANet-inspired and APOLLO-inspired adaptations are labeled
explicitly to distinguish them from the original systems.

\paragraph{Comparison conditions.}
We distinguish task-level baseline comparisons from
component ablations. Baseline comparisons evaluate the
implemented methods, including their respective learning
and inference procedures. Component ablations use matched
experience and task splits to examine individual design
choices in \method. Acquisition, training, and serving costs
are accounted for separately; using the same backbone does
not imply equal computational cost.
\section{Evaluation and Verification Protocol}
\label{app:protocol}

\paragraph{Data partitions and leakage control.}
We distinguish experience used for acquisition and training,
development data used for controller verification, and
held-out data used for final evaluation. The intended unit
of separation is the originating episode: segments from the
same episode should remain in one partition even when they
are associated with different skills. For QA, normalized
questions are deduplicated across partitions before answers
are attached. Question--answer matching provides an additional
check rather than replacing question-level separation.
The final evaluation pool is excluded from gate tuning,
checkpoint selection, and prompt development.

\paragraph{Sequential probe construction.}
In ALFWorld, the consolidator groups identical
observation--action--prompt trajectories across skill labels,
preserves the reserved groups across rounds, and excludes
trajectories already present in replay. Other sequential
environments reserve a deterministic tail of successful
memory records per skill after sorting by node identifier.
Reserved records are excluded from the current consolidation
batch. Observation--action probes include up to 16 recent
history entries and are subject to a probe-step cap.

\paragraph{Verification conditions.}
The candidate arm uses the adapted model without episodic
retrieval; the reference arm uses the base model with ETG.
Both arms use the same task interface and scoring procedure.
The implemented reference arm retrieves from the pre-release
store. Removing a record from the training batch does not
necessarily remove it from this retrieval store, so this
condition is distinct from source-excluded retrieval.

For sequential tasks, probes measure agreement with recorded
actions using the same environment prompt builder and action
parser. Agreement is a local behavioral diagnostic: it can
depend on the action classes represented in the probe and
does not capture all valid alternative actions or subsequent
environment states.

For the QA release analysis in
Figure~\ref{fig:verification-release}, each question is counted
as successful when answer F1 is at least 0.5. The skill-level
score is the fraction of successful development questions.
Release requires this fraction to exceed 50\%, together with
the relative-performance and training-evidence checks.
These answer-level scores are distinct from next-action
agreement.

\paragraph{Statistical unit.}
Paired comparisons match methods by environment world or
task instance. Bootstrap draws resample the same instances
for all compared methods and recompute the reported metric.
For Crafter, each sampled world retains its complete
achievement vector; progress or geometric score is then
computed from the resulting achievement rates.

Paired differences are summarized with percentile 95\%
bootstrap intervals. For evaluations with multiple evolution
seeds, seed-level variation is distinguished from
instance-level uncertainty, with hierarchical resampling
where applicable. Single-checkpoint analyses quantify
variation over evaluation instances rather than variation
across independently trained models. The fraction of
bootstrap draws favoring a method is descriptive and is
not reported as a hypothesis-test $p$-value.

\paragraph{Interaction and compute accounting.}
We distinguish student acquisition, teacher acquisition,
verification, consolidation, and final evaluation costs.
Serving measurements include all model calls and tokens
used by each method, including reflection and repeated
generation. Last-attempt success and best-of-attempt success
are different outcomes and are identified separately when
reported. Matched-data controls reuse the same acquired
experience; this does not imply equal end-to-end acquisition
or training cost.

\paragraph{Decision and memory accounting.}
Release decisions are associated with the evaluated
checkpoint and the supporting training and probe evidence.
The relevant quantities include student mastery, available
teacher demonstrations, realized training examples, probe
size, candidate and reference scores, and the configured
relative threshold and absolute floor.

Memory accounting distinguishes records available for
retrieval, records moved to a non-retrievable archive, and
separately retained training data. Counts before and after
release are reported for the affected stores. Archival
release reduces retrieval availability without necessarily
reducing total storage.

\paragraph{LLM-only control.}
For QA, LLM-only denotes one greedy zero-shot chain-of-thought
generation from the unadapted student using the question and
the benchmark-supplied context, without episodic retrieval,
additional search, or an adapter. Interactive controls with
handcrafted strategy hints or action substitutions are
identified separately rather than labeled LLM-only.

\subsection{Controlled evaluation of memory release}
\label{app:release-controls}

\paragraph{Shared experimental conditions.}
We use the fixed Qwen3-8B SFT checkpoint and the initial
store of 600 HotpotQA experience records analyzed in
Figure~\ref{fig:verification-release}.
All conditions share the checkpoint, skill assignments,
prompt construction, retrieval settings, and decoding
configuration. No additional parameter updates are
performed during this comparison.
Development probes determine release decisions before
held-out evaluation; test results are not used to revise
these decisions or their thresholds.

\paragraph{Release-policy controls.}
We compare the following policies:
\begin{itemize}
    \item \textbf{Retain all.}
    All experience records remain retrievable, and all
    skills use the fixed adapted model with ETG.

    \item \textbf{Immediate release.}
    Skills that contributed valid training examples
    release their eligible records after training,
    without post-training behavioral verification.

    \item \textbf{Verified release.}
    Eligible records are released only for skills that
    satisfy the training-evidence checks, relative
    verification threshold, and absolute performance
    floor.

    \item \textbf{Random release.}
    Release selections are sampled without using probe
    scores and matched to the verified policy's
    retrievable record count.
    We repeat the random selection to measure variation
    due to the choice of released memories.
\end{itemize}

For every policy, skills selected for release use the
same adapted model without episodic retrieval.
Other skills use that model with the remaining ETG.
Adapter activation is fixed rather than independently
recomputed for each policy.
Record eligibility, shared-record handling, and any
configured residual retention are applied consistently
across conditions.
Released records are moved to a non-retrievable archive;
the comparison therefore measures retrieval reduction
rather than total-storage reduction.

\paragraph{Held-out performance and uncertainty.}
We evaluate every condition on the same held-out questions
and report aggregate answer EM and token F1.
For each release policy, we compute the paired F1 difference
relative to retaining all records.
Paired bootstrap draws resample test questions using the
same indices for both conditions and recompute the score
difference, yielding percentile 95\% confidence intervals.

For random release, we report performance averaged over
the sampled release selections and account for both
selection-level and question-level variation through
hierarchical bootstrap resampling.
These intervals quantify uncertainty for the fixed
checkpoint, rather than variation across training seeds.
\section{Implementation and Reproducibility}
\label{app:repro}

Table~\ref{tab:qwen-hyperparameters} and~\ref{tab:llama-hyperparameters} listed hyperparameters used for experiments using Qwen3-8b and llama3.1-8b-instruct.

\begin{table}[t]
\centering\footnotesize\setlength{\tabcolsep}{4pt}
\caption{Benchmark-specific SFT settings for Qwen3-8B in the scaled evaluations. Training examples count decisions for ALFWorld and WebShop, answers for WebQuestions, and decision windows from 32 training worlds for Crafter.}
\label{tab:qwen-hyperparameters}
\begin{tabular}{@{}lrrrr@{}}
\toprule
Setting & ALFWorld & WebShop & WebQuestions & Crafter\\
\midrule
Maximum training tokens & 16,384 & 8,192 & 8,192 & 8,192\\
Data passes & 3 & 3 & 3 & 3\\
Optimizer updates & 5,004 & 2,149 & 450 & 1,281\\
Training seed & 42 & 42 & 42 & 42\\
Learning rate & $2\times10^{-5}$ & $2\times10^{-5}$ & $2\times10^{-5}$ & $2\times10^{-5}$\\
LoRA rank / scaling & 16 / 32 & 16 / 32 & 16 / 32 & 16 / 32\\
LoRA dropout & 0.05 & 0.05 & 0.05 & 0.05\\
Microbatch / accumulation & 1 / 4 & 1 / 4 & 1 / 4 & 1 / 4\\
Weight decay & 0.01 & 0.01 & 0.01 & 0.01\\
Warm-up fraction & 0.1 & 0.1 & 0.1 & 0.1\\
Base-model KL coefficient & 0 & 0 & 0 & 0\\
Training weight loading & 4 bit & 4 bit & 4 bit & 4 bit\\
Initial retrieval records & 19 & 43 & 16 & 185\\
\bottomrule
\end{tabular}
\end{table}

\begin{table}[t]
\centering\footnotesize\setlength{\tabcolsep}{4pt}
\caption{Benchmark-specific SFT settings for Llama-3.1-8B-Instruct in the scaled evaluations. Training examples count decisions for ALFWorld and WebShop, answers for WebQuestions, and decision windows from 32 training worlds for Crafter.}
\label{tab:llama-hyperparameters}
\begin{tabular}{@{}lrrrr@{}}
\toprule
Setting & ALFWorld & WebShop & WebQuestions & Crafter\\
\midrule
Maximum training tokens & 16,384 & 8,192 & 8,192 & 8,192\\
Data passes & 3 & 3 & 3 & 3\\
Optimizer updates & 5,004 & 2,149 & 450 & 1,281\\
Training seed & 42 & 42 & 42 & 42\\
Learning rate & $2\times10^{-5}$ & $2\times10^{-5}$ & $2\times10^{-5}$ & $2\times10^{-5}$\\
LoRA rank / scaling & 16 / 32 & 16 / 32 & 16 / 32 & 16 / 32\\
LoRA dropout & 0.05 & 0.05 & 0.05 & 0.05\\
Microbatch / accumulation & 1 / 4 & 1 / 4 & 1 / 4 & 1 / 4\\
Weight decay & 0.01 & 0.01 & 0.01 & 0.01\\
Warm-up fraction & 0.1 & 0.1 & 0.1 & 0.1\\
Base-model KL coefficient & 0 & 0 & 0 & 0\\
Training weight loading & 4 bit & 4 bit & 4 bit & 4 bit\\
Initial retrieval records & 19 & 43 & 16 & 185\\
\bottomrule
\end{tabular}
\end{table}

\paragraph{Serving conditions.}
A verified skill uses its approved adapter with episodic retrieval disabled. Other skills use the base model with retained ETG. Task instructions, current observations, recent interaction history, and shared environment rules remain part of the input in both conditions. Previously approved versions remain available when a later candidate fails verification.

\paragraph{Routing implementation.}
At inference time, the controller assigns the current request or interaction state to a skill using only deployment-visible information and the same skill taxonomy used during ETG construction. The router does not observe gold answers, episode success, future environment states, verification outcomes, or held-out annotations. Given the resulting skill key, the controller loads the most recent approved adapter version for that skill; if no approved version exists, it executes the Base+ETG route.

\paragraph{Controller implementation.}
The consolidator selects skills, constructs training examples and reserved probes, continues the adapter, loads the candidate into the serving engine, and evaluates release eligibility. The environment prompt builder is shared by training and evaluation. Training or loading failure preserves the source records. Skills with no valid training examples or no eligible probe remain external. The sequential gate filters probe states to the corresponding deployment route. Retrieval storage and replay storage are recorded separately.

\section{Limitations}
\label{sec:discussion}
Release verification relies on finite development probes,
whose coverage may not capture every deployment condition.
Our controlled release study evaluates a fixed checkpoint;
extending this analysis to longer sequences of consolidation
and changing task distributions would clarify the durability
of release decisions. Consolidation schedules and verification
thresholds are also configured per benchmark. Adapting these
settings online is a promising direction for broader deployment.
\clearpage
\section{Essential Prompt Interfaces}
\label{app:prompts}
Training and deployment share the benchmark's task interface. The layouts below retain the task fields and output constraints; benchmark-specific environment rules are omitted for readability.

\lstdefinestyle{engprompt}{basicstyle=\ttfamily\footnotesize,breaklines=true,breakatwhitespace=true,columns=fullflexible,keepspaces=true,showstringspaces=false,frame=single,rulecolor=\color{draftgray},framesep=5pt,xleftmargin=6pt,xrightmargin=6pt,aboveskip=8pt,belowskip=8pt}
\lstset{style=engprompt}

\paragraph{Task execution.}
Sequential tasks provide the goal, current observation, valid actions, and recent interaction history. The model returns one executable action. Supervised examples pair this pre-action input with the recorded action; the resulting state supplies outcome evidence.
\begin{lstlisting}
Goal: {task goal}
Observation: {current environment state}
Valid actions: {admissible actions}
Recent history: {observations and executed actions}
Output exactly ONE action on its own line.
\end{lstlisting}

QA uses the same question and evidence in training, verification, and matched evaluation arms. The context is omitted in the question-only control.
\begin{lstlisting}
Answer the following question based on the provided context. Think step by step, then give your final answer after 'Answer:'.

Context: {evidence}
Question: {question}
Let's think step by step:
\end{lstlisting}

\paragraph{Experience acquisition and retrieval.}
Teacher acquisition adds the selected skill and its target outcome to the task brief. Successful demonstrations retain their executed actions and observed outcomes. When ETG retrieval is enabled, the task input also contains relevant successful traces or corrections, including their conditions of use. A correction contributes a supervised action only when it has a valid target for an aligned observation.

\paragraph{Verification.}
The two probe arms receive identical task inputs and targets. The candidate arm uses the updated adapter with episodic retrieval disabled; the reference arm uses the base model with ETG retrieval enabled. Both use the same output parser and benchmark-specific probe criterion. Eq.~\eqref{eq:evict} determines release eligibility from these scores and valid training participation.

\clearpage
\section{ETG Records and Successful Deployment Cases}
\label{app:cases}
We follow two successful deployments from the completed WebShop and HotpotQA evaluations. Each case pairs the held-out result with an acquired ETG record, its training participation, and the skill's verification history. Figure~\ref{fig:etg-records} shows the key fields of the actual records.

\subsection{WebShop: retained experience and approved weights complete one purchase}
\label{app:case-webshop}
\paragraph{Acquisition and consolidation.}
A teacher successfully purchases a kelly-green hoodie in size 3x-large for less than \$60. The ETG associates the trajectory with \emph{option selection} and records its observations, executed actions, teacher source, and reward of 1. The first consolidation update includes this trace among 288 action-training examples. On the reserved option-selection probe, both the candidate without ETG and the base model with ETG match all three eligible target actions. The controller approves the adapter for this skill and releases eligible records, including the illustrated teacher record, while keeping residual experience.

The same update retains query formulation, result selection, and purchase decision. A later candidate receives no new approval; deployment continues using the previously verified version for approved skills. This preserves useful weights and external traces across updates.

\paragraph{Successful held-out execution.}
A different goal requests a \emph{navy-blue} hoodie in size 3x-large for less than \$60. This exact goal is absent from the acquired graph snapshots. Table~\ref{tab:case-webshop-actions} shows the five executed actions. The initial search uses the base model with retained ETG guidance; the remaining four actions use the approved adapter with episodic retrieval disabled.

\begin{table}[h]
\centering\small\setlength{\tabcolsep}{5pt}
\caption{WebShop successful trajectory in a separate completed run. The search query is shortened for display; product identifiers and subsequent actions are preserved.}
\label{tab:case-webshop-actions}
\begin{tabular}{rll}
\toprule
Step & Executed action & Policy / experience access\\
\midrule
1 & \texttt{search[...navy blue 3x-large]} & Base + retained ETG\\
2 & \texttt{click[b01hqtwl6s]} & Approved adapter; ETG off\\
3 & \texttt{click[navy blue]} & Approved adapter; ETG off\\
4 & \texttt{click[3x-large]} & Approved adapter; ETG off\\
5 & \texttt{click[buy now]} & Approved adapter; ETG off\\
\bottomrule
\end{tabular}
\end{table}

The environment awards full purchase reward after five actions. Under the same goal and interface, Base repeatedly searches and inspects unsuitable products, exhausting 30 actions with reward 0. This separate 100-goal evaluation obtains 47 successes with \method\ and 44 with Base: 14 tasks change from failure to success, and 11 change in the opposite direction. This case exhibits the combined deployment path, with external guidance supporting the query and verified weights handling the subsequent interaction.

\subsection{HotpotQA: a retained bridge trace is later released after verification}
\label{app:case-hotpotqa}
\paragraph{An ETG record with a reusable reasoning pattern.}
A successful teacher example asks which actor appeared in both \emph{D.C. Cab} and \emph{Barney Miller}. The stored trace identifies Max Gail in the film cast and connects him to his television role. The record carries the teacher source tag, supplied evidence, answer, and \emph{bridge reasoning} skill association.

The record contributes valid training data to both consolidation updates. After the first update, the weights arm passes 2 of 5 reserved bridge-reasoning probes and the memory arm passes 4 of 5; the record remains in ETG. After the second update, the scores become 3 of 5 and 2 of 5. The controller approves that skill's new adapter version and releases the illustrated record. Replay remains available for later training. Thus the record's release follows verified behavior after a later update, rather than the completion of its first training exposure.

\paragraph{Successful reasoning without episodic retrieval.}
The held-out question asks:
\begin{quote}\small
In which University is Richard Williams' former assistant a member of the Athletics Hall of Fame?
\end{quote}
Both models receive the same ten supplied paragraphs. The relevant evidence identifies Rick Stansbury as Williams' former assistant and states that Stansbury belongs to the Campbellsville University Athletics Hall of Fame. The base model answers \emph{Mississippi State}, the university where Williams coached. The approved adapter answers \emph{Campbellsville University}, matching the reference answer.

The question is absent from the acquired ETG and reserved-probe records. The evaluated request uses the second approved adapter with no episodic retrieval. Its response follows the intermediate person to the requested affiliation, while Base selects the university associated with the first named person. The acquisition trace and this evaluation question involve different entities but share the bridge-reasoning skill.

\end{document}